\documentclass[conference]{IEEEtran}
\IEEEoverridecommandlockouts
\usepackage{cite}
\usepackage{amsmath,amssymb,amsfonts}
\usepackage{algorithmic}
\usepackage{graphicx}
\usepackage{textcomp}
\usepackage{xcolor}
\usepackage{flushend}
\usepackage{enumitem, url}
\def\BibTeX{{\rm B\kern-.05em{\sc i\kern-.025em b}\kern-.08em
    T\kern-.1667em\lower.7ex\hbox{E}\kern-.125emX}}
\begin{document}

\title{Object Recognition in Atmospheric Turbulence Scenes\\

\thanks{This work was supported by the UKRI MyWorld Strength in Places Programme (SIPF00006/1).}
}

\author{\IEEEauthorblockN{Disen Hu}
\IEEEauthorblockA{\textit{Visual Information Laboratory} \\
\textit{University of Bristol}\\
Bristol, UK \\
zr18100@bristol.ac.uk}
\and
\IEEEauthorblockN{ Nantheera Anantrasirichai }
\IEEEauthorblockA{\textit{Visual Information Laboratory} \\
\textit{University of Bristol}\\
Bristol, UK \\
n.anantrasirichai@bristol.ac.uk}
}

\maketitle

\begin{abstract}
The influence of atmospheric turbulence on acquired surveillance imagery poses significant challenges in image interpretation and scene analysis. Conventional approaches for target classification and tracking are less effective under such conditions. While deep-learning-based object detection methods have shown great success in normal conditions, they cannot be directly applied to atmospheric turbulence sequences. In this paper, we propose a novel framework that learns distorted features to detect and classify object types in turbulent environments. Specifically, we utilise deformable convolutions to handle spatial turbulent displacement. Features are extracted using a feature pyramid network, and Faster R-CNN is employed as the object detector. Experimental results on a synthetic VOC dataset demonstrate that the proposed framework outperforms the benchmark with a mean Average Precision (mAP) score exceeding 30\%. Additionally, subjective results on real data show significant improvement in performance.
\end{abstract}

\begin{IEEEkeywords}
atmospheric turbulence, object detection, deep learning, deformable convolution, object recognition
\end{IEEEkeywords}

\section{Introduction}
Atmospheric turbulence consistently degrades visual quality and has a negative impact on the performance of automated target recognition and tracking in a scene. These distortions occur when there is a temperature difference between the ground and the air, causing rapid upward movement of air layers and resulting in changes to the interference pattern of light refraction. This leads to visible ripples and waves in both spatial and temporal directions in images and videos.
Mitigating this effect is an ill-posed problem with non-stationary distortions varying across time and space, and a degree of distortions is unknown. The restoration process is thus complex and time-consuming. Objects behind the distorting layers are almost impossible to recognise by machines leading to a failure of automatic detection and tracking processes. Examples of applications directly affected with atmospheric turbulence are video surveillance, security and defence.

Object detection methods on natural and clean images have been greatly developed achieving high performance in term of both detection accuracy and computational speed. The state of the arts are based on deep learning with convolutional neural networks (CNN) (see recent techniques for object detection in \cite{Anantrasirichai:Artificial:2021}). However, the performance of these methods declines when the features are corrupted by noise or distorted by blur. This degradation is even more pronounced in the case of atmospheric turbulence, where distortions appear randomly and are spatially and temporally invariant \cite{Deshmukh:moving:2013}. Up to now, only face recognition in atmospheric turbulence has been developed \cite{Lau:ATFaceGAN:2020}. They have also proved that training deblurring and detection models together gives better results than separating the models.

In this paper, we tackle the multi-class object detection in atmospheric turbulence scenes without image restoration process. Our framework is hence fast and straightforward. The method is developed based on the Faster R-CNN detector \cite{Ren:FasterRCNN:2017}, but this should not be limited to. The features are extracted with a feature pyramid network (FPN) \cite{Lin:Feature:2017}, which can deal with different resolutions, different sizes of the objects, and different amounts of distortions.   To mitigate the effects of atmospheric turbulence, we incorporate deformable convolutions \cite{Dai:Deformable:2017}, which help reduce the impact of visible ripples along object edges caused by atmospheric turbulence. 
A key contribution is that, as in the atmospheric turbulent environments, the objects exhibit visual distortions within small ranges of pixel displacement appearing randomly at all directions. The use of deformable convolutions provides flexibility in capturing the shapes of the objects and assists the FPN in extracting the appropriate features from the distorted objects. As there is no ground truth available for this problem, we trained the model using a synthetic dataset and evaluated its performance using both synthetic and real datasets. Our code is available at \url{https://github.com/disen-hu/FasterRcnn_FPN_DCN}.

\section{Related work}
\label{sec:related}

Image restoration techniques for atmospheric turbulence have been extensively studied \cite{Anantrasirichai:Mitigating:2012, Anantrasirichai:Atmospheric:2013, Patel:adaptive:2019}, with some methods specifically addressing moving objects in distorted scenes \cite{Deshmukh:moving:2013, Halder:geometric:2015, Anantrasirichai:Atmospheric:2018, Mao:Image:2020}. Deep learning technologies have also gained attention in atmospheric turbulence mitigation, although they are still in the early stages of development. Existing architectures have been employed and retrained using synthetic datasets in various methods \cite{Gao:Atmospheric:2019, Mao:accelaring:2021, Chak:Subsampled:2021}. Additionally, the use of Complex-Valued CNN was explored in \cite{Anantrasirichai:Atmospheric:2022}, demonstrating significant improvements over traditional CNN-based approaches.

For object detection in atmospheric turbulence scenes, most traditional methods were proposed to detect long-distance target objects. The main limitation of these is that those objects are at sufficiently low fidelity to exhibit little or no detail, instead appearing as blurred silhouettes or blobs \cite{Deshmukh:moving:2013, Oreifej:Simultaneous:2013, Chen:detecting:2014, Gilles:Detection:2018, Zhang:Stabilization:2018 }. To the best of our knowledge, the deep learning-based object detection has been applied for only face recognition \cite{Lau:ATFaceGAN:2020}. This method is based on Generative Adversarial Networks (GAN). However, the features on human faces are very clear and distinguish from surroundings. This method is therefore not suitable for complicated scenes with many types of objects. For general object recognition, there is not any deep learning methods specifically proposed for that in atmospheric turbulent environment.
Authors in \cite{Uzun:Augmentation:2022} has tested three state-of-the-art object detection methods, retrained with the synthetic thermal imagery. They reported that among VfNet \cite{Zhang:VarifocalNet:2021}, YOLOR \cite{Wang:YOLOR:2021}, and TOOD \cite{Feng:TOOD:2021}, YOLOR gave the best performance in both mean average precision and speed.

\section{Methodology}
\label{sec:method}

\subsection{Dealing with atmospheric turbulent distortion}
\label{ssec:dealing}
A turbulent medium causes phase fluctuations, which exhibit in the image as a phase shift and its amount depends approximately linearly on spatial displacement \cite{Hill:Undecimated:2015}. Following quasi-periodic property, the phase of each pixel is altered randomly, whilst the magnitude of high frequency is generally decreased due to mixing of the signals leading to a blur. This causes images look like ripples across time -- the pixels spatially shift from their actual locations in random directions. In term of image degradation model, the distorted image $I_\text{atmos}$ is described as $I_\text{atmos} = h*I_\text{ideal} + n$, where $n$ is noise, and $h$ is an unknown spatially variant point spread function, comprising geometric distortion and blur. 

The clean or ideal image $I_\text{ideal}$ is altered by $h$ more than $n$ significantly.  That is, there are two aspects of atmospheric turbulence to concern: i) ripple effect (geometric distortion and blur) and ii) unknown amount of this effect locally. In this paper, We exploit deformable convolutions to deal with the ripple effect. Mathematically, the deformable convolution can be expressed as Eq. \ref{eq:deformconv}. Following \cite{Dai:Deformable:2017}, the output feature map $y$ at pixel $p$ is the result of convolution between a learnable weight $w(p_n)$, where $p_n$ are elements of regular grid $R$. The irregular positions of an input feature map $x$ are a combination of the regular grid $R$ and the offsets $\Delta p_n$.

\begin{equation}
    y(p) = \sum_{p_n \in R} w(p_n) \cdot x(p + p_n + \Delta p_n)
    \label{eq:deformconv}
\end{equation}

\begin{figure} [t!]
    \centering
    \includegraphics[width=0.9\columnwidth]{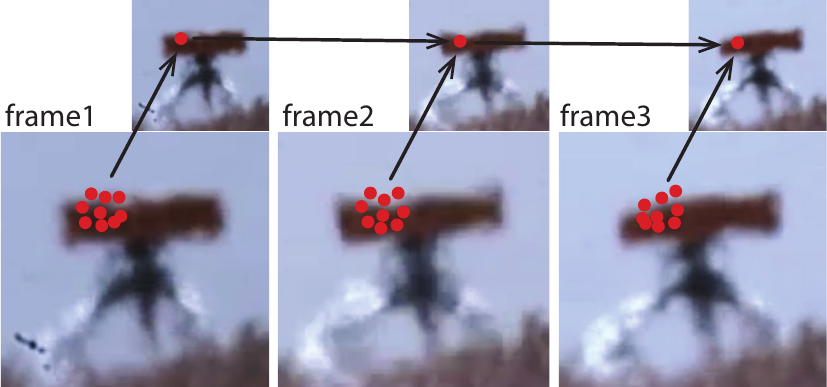}
    \caption{Visualisation of the effect of 3$\times$3 deformable convolutions on turbulence distortion. The red dots correspond to the same feature. }
    \label{fig:deform}
\end{figure}

\noindent This can be interpreted that the deformable convolutions allow the pixels associated to the current kernel to spatially locate outside the regular grid search. So the same object in the different images, affected by different turbulent distortions, could then have the same features, as shown in Fig. \ref{fig:deform}. This fully benefits in object recognition as gaining more accurate features.

To deal with unknown amount of distortions, we exploit multiple scales of features so that they cover all possible ranges of displacement duo to pixels  in $I_\text{atmos}$ shifted from those in $I_\text{ideal}$.

\subsection{Network architectures}

\begin{figure} [t!]
    \centering
    \includegraphics[width=\columnwidth]{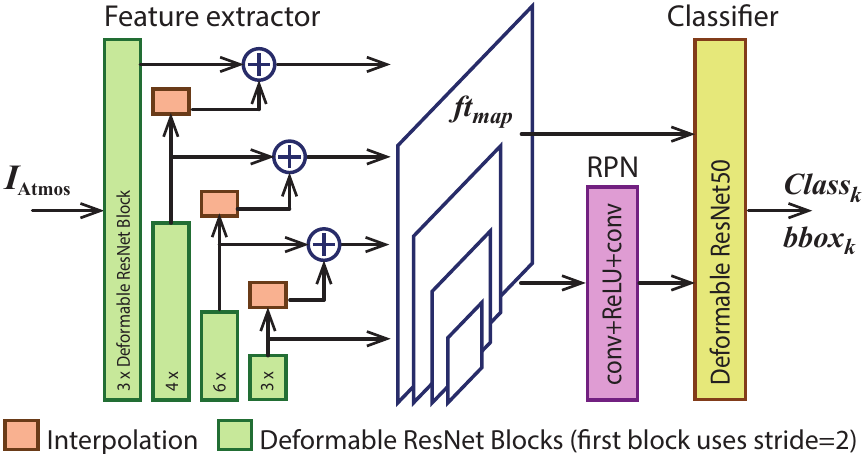}
    \caption{A diagram of the proposed object detection in atmospheric turbulence scenes. }
    \label{fig:diagram}
\end{figure}

The proposed framework adapted from Faster R-CNN \cite{Ren:FasterRCNN:2017} is shown in Fig. \ref{fig:diagram}, comprising three modules: i) feature extraction, ii) region proposal network (RPN) and iii) classification network. 

For the feature extraction, we replace the single-scale feature extractor originally used in the Faster R-CNN with a feature pyramid network (FPN) \cite{Lin:Feature:2017}.  FPN offers multi-scale features which benefit i) different sizes of targeting objects, ii) deal with different amount of atmospheric turbulence, and iii) improve classification accuracy. We initial the FPN with pretrained model on COCO segmentation dataset \cite{Lin:COCO:2014}.
We tested two backbone architectures, VGG16 \cite{Simonyan:VGG:2015} and ResNet50 \cite{He:ResNet:2016}, and found the ResNet50-based model outperforms the VGG16-based model more than double in term of mAP. Therefore, we employ ResNet50.
We then replace regular grid convolutions with deformable convolutions following the reasons stated in Section \ref{ssec:dealing}. 

RPN generates bounding box proposals in different sizes using shallow architecture (3$\times$3 Conv + ReLU + 1$\times$1 Conv). The region proposals are sent to the RoI Pooling layer together with feature maps for classification. We employ pretrained ResNet50 model from the official PyTorch, which has achieved 81\% accuracy on VOC datasets. Using a pretrained model can reduce training duration and achieve better results. Similar to the feature extractor, we apply deformable convolutions in RPN and in the ResNet blocks. At the end, the feature vectors are passed to two parallel fully connected layers. One branch uses Softmax function to calculate the probability of being each category, and the other branch outputs the coordinate offset of the proposal box and uses the frame regression to correct the coordinate. Non-maximum Suppression (NMS) is employed to eliminate the redundant boxes.

\subsection{Training procedure}

Since the Faster R-CNN combines both classification and bounding box regression, the loss function $L$ of the proposed framework consists of two parts: RPN loss and regression loss of Faster R-CNN as shown in the first and the second parts of Eq. \ref{eq:loss}, respectively.

\begin{equation}
\label{eq:loss}
\begin{split}
L &= - \frac{1}{N_f} \sum\nolimits_{i=1}^{N_f} \text{log} [p_i^* p_i+(1-p_i^*)(1-p_i)] \\
& \ \ + \alpha \frac{1}{N_r} \sum\nolimits_{i=1}^{N_r} p_i^* \text{smooth}_{L_1} (t_i - t_i^*),
\end{split}
\end{equation}

\noindent where $p_i$ represents a probability that anchor forecast is the object $i$, $p^*$ is equal 0 or 1 when ground truth is negative or positive, respectively. $t_i$ represents the prediction in RPN training stage, of which the ground truth is $t^*$. The smooth L1 loss is defined as Eq. \ref{eq:smoothL1}, where $\sigma$ is empirically set to 3.

\begin{equation}
\label{eq:smoothL1}
\text{smooth}_{L_1} (x) = \begin{cases}
0.5 x^2 \frac{1}{\sigma^2}, \ \text{if} \  |x| < \frac{1}{\sigma^2} \\
|x| - 0.5, \ \text{otherwise}
\end{cases}
\end{equation}

\noindent We employ Adam optimizer with an initial learning rate of 0.0001.
Three ratios of RPN anchor are used, i.e. 0.5 , 1 and 2. The IoU thresholds of RPN and RCNN for positive detection are set to  0.7 and 0.5, respectively.

\section{Results and discussion}
\label{sec:results}

Our framework was trained with synthetic dataset, generated following the procedure in \cite{Gao:Atmospheric:2019}, where nine point spread functions of atmospheric turbulence were locally applied with random size and strength. We used VOC dataset \cite{Everingham:voc:2015}, containing 20 classes\footnote{VOC dataset inclues person, bird, cat, cow, dog, horse, sheep, aeroplane, bicycle, boat, bus, car, motorbike, train, bottle, chair, dining table, potted plant, sofa, tv/monitor} labelled as bounding boxes. There are 16,551 images for training and 4,952 images for testing, with total 52,090 annotated objects. The results of this synthetic testing dataset were presented in Section \ref{ssec:syn}, and the inference of training with this synthetic training dataset was then used to detect objects in the real dataset, as shown in Section \ref{ssec:real}.

\subsection{Synthetic datasets} \label{ssec:syn}
We compared our methods with state-of-the-are object detection, YOLOv4 \cite{Bochkovskiy:YOLOv4:2020}. YOLOv4 employs Path Aggregation Network (PAN) which is the same concept of FPN, so we only implemented deformable convolutions on YOLOv4 framework (YOLOv4+DC). The results are shown in Table \ref{tab:results}, and our proposed method (Faster R-CNN+FPN+DC) achieves the best performance in all three metrics used. 
Both Faster R-CNN and YOLOv4 are very successful in common object detection datasets. However, for the object detection problem with atmospheric turbulence effect, the background is blurred, and objects are distorted due to atmospheric turbulence, which makes them insufficient to solve this problem.

\begin{figure}[t!]
    \centering
    \includegraphics[width=\columnwidth]{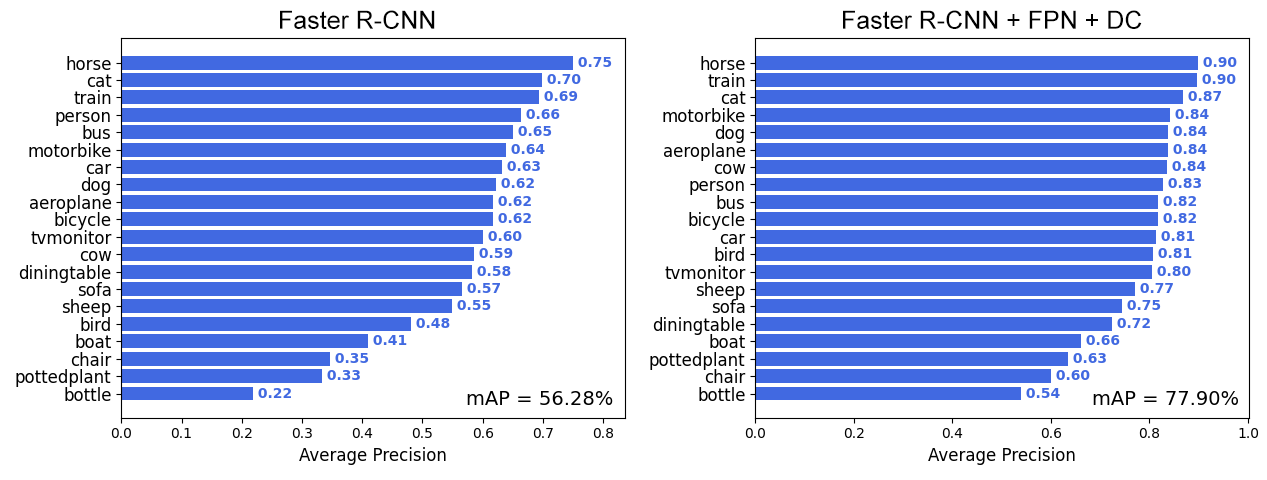}
    \caption{An improvement of mAP in each category when FPN and deformable convolutions (DC) are employed.}
    \label{fig:mAPs}
    \end{figure}

\begin{figure*}[t!]
    \centering
    \includegraphics[width=\textwidth]{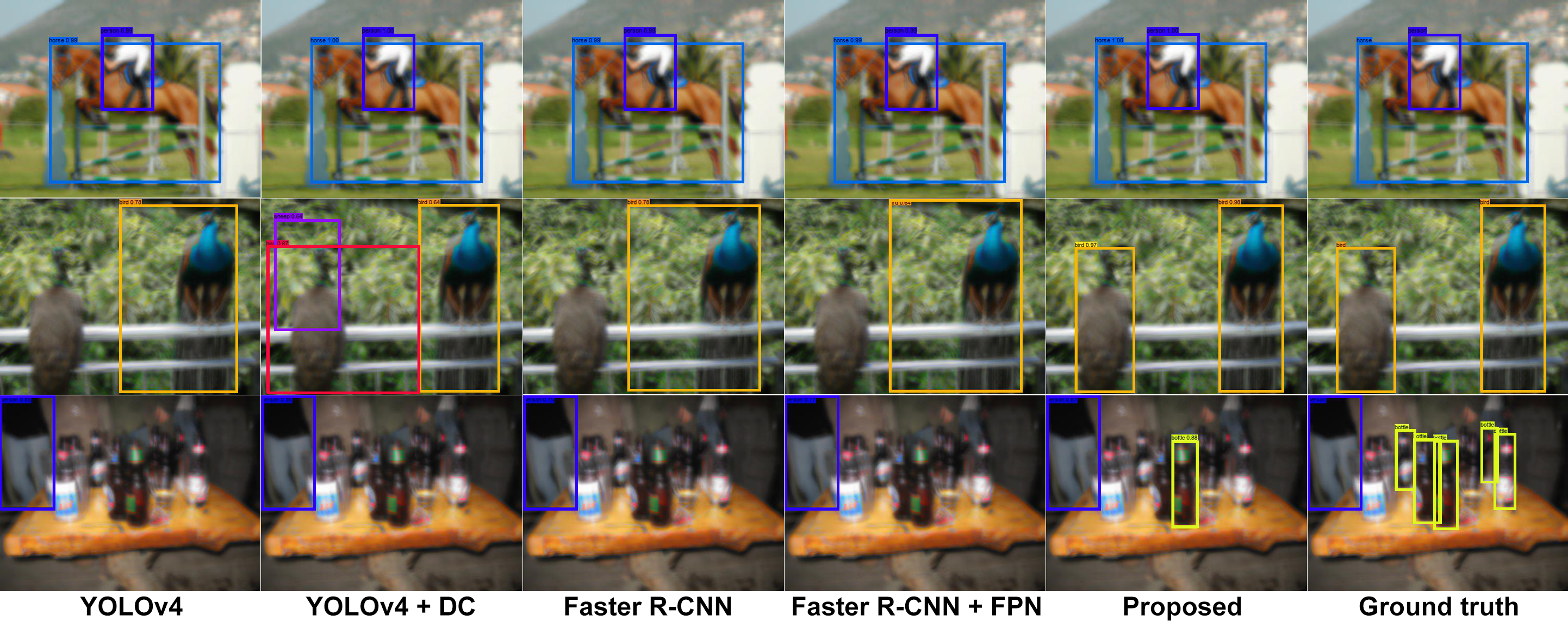}
    \caption{Subjective results of different models and ground truth. The top to bottom rows show the easy, medium and difficult cases, respectively.}
    \label{fig:resultcompare}
\end{figure*}

 We examined the influence of FPN and deformable convolutions. Fig. \ref{fig:mAPs} reveals a significant improvement on the detection performance. The highest precision was achieved when detecting horses. Detecting the bottles is the most improvement -- FPN alone improved mAP by 55\%, and FPN and DC together improved as high as 145\%. The most missed objects contain thin structures, e.g. bottles, chair, and boat, as these parts could be distorted, blurred and their features are mixed with the background. The subjective results in Fig. \ref{fig:resultcompare} show examples of i) easy case, where all models can detect the house and the person correctly; ii) medium case, where only the proposed method achieves accurate results, whilst others can only detect the right bird correctly; iii) the hard case, where all models can detect the person but most models missed out the bottles.
 
\begin{table}[t!]
    \centering
    \caption{Detection results on the synthetic datasets with mean Average Precision (mAP), mean Average Recall (mAR) and f1-score }
\begin{tabular}{ l c c c}
 \hline
 Method & mAP & mAR & f1-score \\ 
 \hline
 YOLOv4 & 0.598 & 0.610 & 0.562  \\  
 YOLOv4 + DC & 0.654 & 0.672 &  0.609   \\
 Faster R-CNN & 0.563 & 0.576 &  0.542 \\
 Faster R-CNN + FPN & 0.630 & 0.650 & 0.583 \\
 Faster R-CNN + FPN + DC & \textbf{0.779} & \textbf{0.773} & \textbf{0.703} \\
  \hline
\end{tabular}
    \label{tab:results}
\end{table}

\begin{figure*}[t!]
    \centering
        \includegraphics[width=\textwidth]{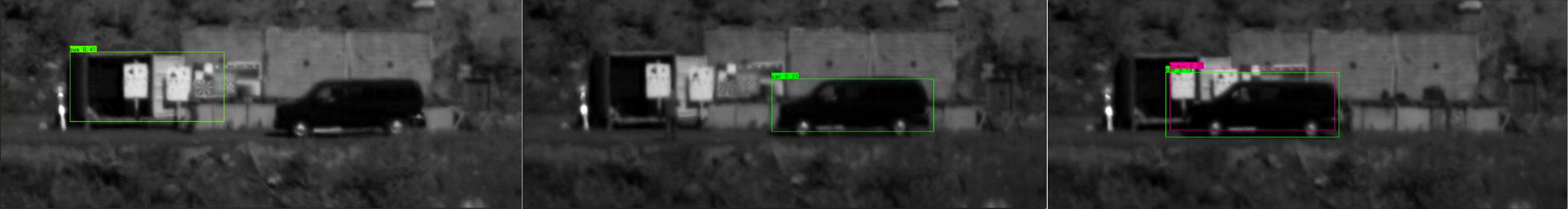}
    \caption{Subjective results of 1100$\times$440 `Truck' sequence. (Left-Right) The incorrect detection, correct detected, mixed results, respectively.}
    \label{fig:blackandwhitecar}
\end{figure*}

\subsection{Real datasets} \label{ssec:real}
\begin{figure}[t!]
    \centering
    \includegraphics[width=\columnwidth]{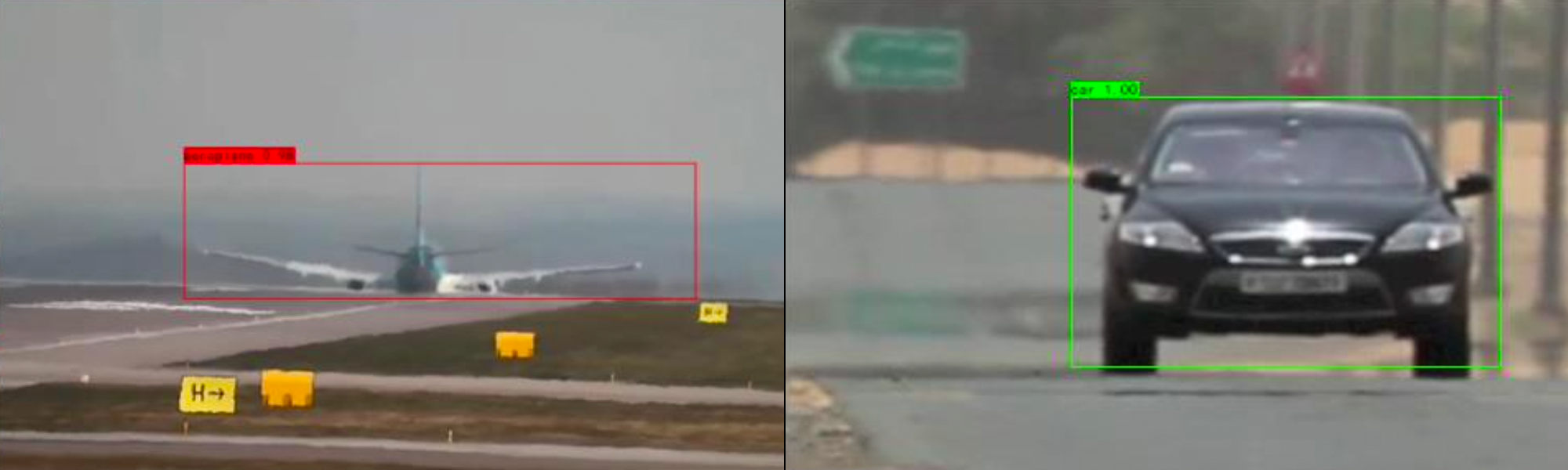}
    \caption{Detection results of 640$\times$360 `Airport' and `Car' sequences. }
    \label{fig:planeandcar}
\end{figure}

\begin{figure}[t!]
\centering
    \includegraphics[width=\columnwidth]{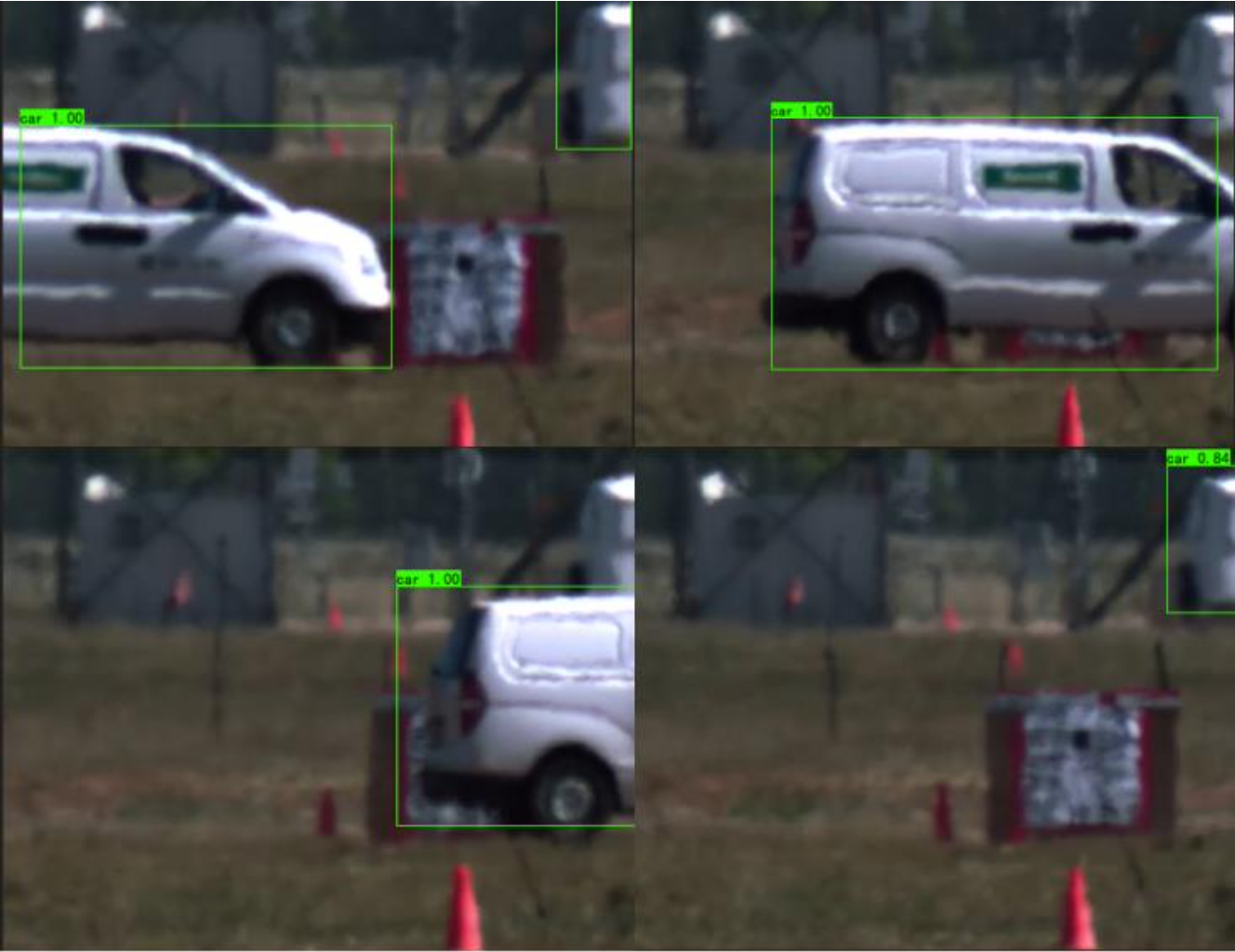}
    \caption{Detection results of 480$\times$384 `Van' sequence. }
    \label{fig:van}
\end{figure}

We explored the performance of our proposed model on real atmospheric turbulence datasets, detailed in \cite{Anantrasirichai:Atmospheric:2022}. The objective assessment was however not applicable to the real distortion datasets due to the absence of ground truth. Fig. \ref{fig:planeandcar} shows that our method correctly detected the aeroplane and the car with the probabilities of 0.89 and 1.0, respectively.  Fig. \ref{fig:van} shows the frame-by-frame detection of a `Van' sequence. The vans in the top row and the bottom-left frames were detected as a car, but the bottom-right frame shows the incorrect detection. Notably, despite being trained on synthetic datasets, as described above, the model detected a white building, visually similar to a part of the van, as a car. Fig. \ref{fig:blackandwhitecar} presents a challenging sequence example. A group of objects, including a container, several rectangular sign posts, in many frames was consistently detected as a car as shown in Fig. \ref{fig:blackandwhitecar} left. However, the probabilities were not high ($<$0.5). Fig. \ref{fig:blackandwhitecar}  right shows that the van was detected as a car and also a train with probabilities of 0.74 and 0.33, respectively. The incorrect detection may be attributed to the lack of color information in this sequence and the strong distortion effect.


\section{Conclusions}
\label{sec:conclusion}

This paper presents a novel framework for object recognition in the atmospheric turbulence environment, where image quality is severely degraded due to random signal interference. The framework is based on Faster R-CNN and utilizes deformable convolutions to handle spatial displacement caused by turbulence. The feature extraction is performed in a pyramid manner, considering different object sizes and levels of distortion. Experimental results demonstrate that the proposed method outperforms state-of-the-art object detection methods significantly.

\bibliographystyle{IEEEtran}
\bibliography{ref_heathaze}

\begin{thebibliography}{10}

\bibitem{Anantrasirichai:Artificial:2021}
N.~Anantrasirichai and David Bull,
\newblock ``Artificial intelligence in the creative industries: a review,''
\newblock {\em Artifcial Intelligence Review}, 2021.

\bibitem{Deshmukh:moving:2013}
A.~S. Deshmukh, S.~S. Medasani, and G.~R. Reddy,
\newblock ``Moving object detection from images distorted by atmospheric
  turbulence,''
\newblock in {\em International Conference on Intelligent Systems and Signal
  Processing}, March 2013, pp. 122--127.

\bibitem{Lau:ATFaceGAN:2020}
Chun~Pong Lau, Hossein Souri, and Rama Chellappa,
\newblock ``{ATFaceGAN: S}ingle face image restoration and recognition from
  atmospheric turbulence,''
\newblock in {\em IEEE International Conference on Automatic Face and Gesture
  Recognition}, 2020, pp. 32--39.

\bibitem{Ren:FasterRCNN:2017}
S.~{Ren}, K.~{He}, R.~{Girshick}, and J.~{Sun},
\newblock ``{Faster R-CNN}: {T}owards real-time object detection with region
  proposal networks,''
\newblock {\em IEEE Transactions on Pattern Analysis and Machine Intelligence},
  vol. 39, no. 6, pp. 1137--1149, 2017.

\bibitem{Lin:Feature:2017}
T.~{Lin}, P.~{Dollár}, R.~{Girshick}, K.~{He}, B.~{Hariharan}, and
  S.~{Belongie},
\newblock ``Feature pyramid networks for object detection,''
\newblock in {\em CVPR}, 2017, pp. 936--944.

\bibitem{Dai:Deformable:2017}
J.~{Dai}, H.~{Qi}, Y.~{Xiong}, Y.~{Li}, G.~{Zhang}, H.~{Hu}, and Y.~{Wei},
\newblock ``Deformable convolutional networks,''
\newblock in {\em ICCV}, Oct 2017, pp. 764--773.

\bibitem{Anantrasirichai:Mitigating:2012}
N.~Anantrasirichai, Alin Achim, David Bull, and Nick Kingsbury,
\newblock ``Mitigating the effects of atmospheric distortion using dt-cwt
  fusion,''
\newblock in {\em ICIP}, 2012, pp. 3033--3036.

\bibitem{Anantrasirichai:Atmospheric:2013}
N.~Anantrasirichai, A.~Achim, N.G. Kingsbury, and D.R. Bull,
\newblock ``Atmospheric turbulence mitigation using complex wavelet-based
  fusion,''
\newblock {\em Image Processing, IEEE Transactions on}, vol. 22, no. 6, pp.
  2398--2408, 2013.

\bibitem{Patel:adaptive:2019}
Akshay Patel, Dippal Israni, Nerella~Arun Mani~Kumar, and Chintan Bhatt,
\newblock ``An adaptive image registration technique to remove atmospheric
  turbulence,''
\newblock {\em Statistics, Optimization and Information Computing}, vol. 7, no.
  2, pp. 439--446, May 2019.

\bibitem{Halder:geometric:2015}
Kalyan~Kumar Halder, Murat Tahtali, and Sreenatha~G. Anavatti,
\newblock ``Geometric correction of atmospheric turbulence-degraded video
  containing moving objects,''
\newblock {\em Optics Express}, vol. 23, pp. 5091--5101, 2015.

\bibitem{Anantrasirichai:Atmospheric:2018}
N.~{Anantrasirichai}, A.~{Achim}, and D.~{Bull},
\newblock ``Atmospheric turbulence mitigation for sequences with moving objects
  using recursive image fusion,''
\newblock in {\em ICIP}, 2018, pp. 2895--2899.

\bibitem{Mao:Image:2020}
Zhiyuan Mao, Nicholas Chimitt, and Stanley~H. Chan,
\newblock ``Image reconstruction of static and dynamic scenes through
  anisoplanatic turbulence,''
\newblock {\em IEEE Transactions on Computational Imaging}, vol. 6, pp.
  1415--1428, 2020.

\bibitem{Gao:Atmospheric:2019}
Jing Gao, N.~Anantrasirichai, and David Bull,
\newblock ``Atmospheric turbulence removal using convolutional neural
  network,''
\newblock in {\em arXiv:1912.11350}, 2019.

\bibitem{Mao:accelaring:2021}
Zhiyuan Mao, Nicholas Chimitt, and Stanley~H. Chan,
\newblock ``Accelerating atmospheric turbulence simulation via learned
  phase-to-space transform,''
\newblock in {\em ICCV}, 2021.

\bibitem{Chak:Subsampled:2021}
Wai~Ho Chak, Chun~Pong Lau, and Lok~Ming Lui,
\newblock ``Subsampled turbulence removal network,''
\newblock {\em Mathematics, Computation and Geometry of Data}, vol. 1, no. 1,
  pp. 1--33, 2021.

\bibitem{Anantrasirichai:Atmospheric:2022}
Nantheera Anantrasirichai,
\newblock ``Atmospheric turbulence removal with complex-valued convolutional
  neural network,''
\newblock {\em arXiv:2204.06989}, 2022.

\bibitem{Oreifej:Simultaneous:2013}
Omar Oreifej, Xin Li, and Mubarak Shah,
\newblock ``Simultaneous video stabilization and moving object detection in
  turbulence,''
\newblock {\em IEEE Transactions on Pattern Analysis and Machine Intelligence},
  vol. 35, no. 2, pp. 450--462, 2013.

\bibitem{Chen:detecting:2014}
Eli Chen, Oren Haik, and Yitzhak Yitzhaky,
\newblock ``Detecting and tracking moving objects in long-distance imaging
  through turbulent medium,''
\newblock {\em Applied Optics}, vol. 53, pp. 1181--1190, 2014.

\bibitem{Gilles:Detection:2018}
J.~Gilles, F.~Alvarez, N.~Ferrante, M.~Fortman, L.~Tahir, A.~Tarter, and
  A.~{von Seeger},
\newblock ``Detection of moving objects through turbulent media. decomposition
  of oscillatory vs non-oscillatory spatio-temporal vector fields,''
\newblock {\em Image and Vision Computing}, vol. 73, pp. 40--55, 2018.

\bibitem{Zhang:Stabilization:2018}
Chao Zhang, Fugen Zhou, Bindang Xue, and Wenfang Xue,
\newblock ``Stabilization of atmospheric turbulence-distorted video containing
  moving objects using the monogenic signal,''
\newblock {\em Signal Processing: Image Communication}, vol. 63, pp. 19--29,
  2018.

\bibitem{Uzun:Augmentation:2022}
Engin Uzun, Ahmet~An{\i}l Dursun, and Erdem Akag\"und\"uz,
\newblock ``Augmentation of atmospheric turbulence effects on thermal adapted
  object detection models,''
\newblock in {\em CVPRW}, June 2022, pp. 241--248.

\bibitem{Zhang:VarifocalNet:2021}
Haoyang Zhang, Ying Wang, Feras Dayoub, and Niko Sünderhauf,
\newblock ``{VarifocalNet: An IoU}-aware dense object detector,''
\newblock in {\em CVPR}, 2021, pp. 8510--8519.

\bibitem{Wang:YOLOR:2021}
Chien-Yao Wang, I-Hau Yeh, and Hong-Yuan~Mark Liao,
\newblock ``You only learn one representation: Unified network for multiple
  tasks,''
\newblock {\em arXiv preprint arXiv:2105.04206}, 2021.

\bibitem{Feng:TOOD:2021}
Chengjian Feng, Yujie Zhong, Yu~Gao, Matthew~R. Scott, and Weilin Huang,
\newblock ``{TOOD: Task-aligned One-stage Object Detection},''
\newblock in {\em ICCV}, 2021, pp. 3490--3499.

\bibitem{Hill:Undecimated:2015}
P.R. Hill, N.~Anantrasirichai, A.~Achim, M.E. Al-Mualla, and D.~Bull,
\newblock ``Undecimated dual tree complex wavelet transforms,''
\newblock {\em Signal Processing: Image Communication}, vol. 35, pp. 61--70,
  2015.

\bibitem{Lin:COCO:2014}
Tsung-Yi Lin, Michael Maire, Serge Belongie, James Hays, Pietro Perona, Deva
  Ramanan, Piotr Doll{\'a}r, and C.~Lawrence Zitnick,
\newblock ``{Microsoft COCO: C}ommon objects in context,''
\newblock in {\em ECCV}, 2014, pp. 740--755.

\bibitem{Simonyan:VGG:2015}
Karen Simonyan and Andrew Zisserman,
\newblock ``Very deep convolutional networks for large-scale image
  recognition,''
\newblock in {\em International Conference on Learning Representations}, 2015.

\bibitem{He:ResNet:2016}
K.~{He}, X.~{Zhang}, S.~{Ren}, and J.~{Sun},
\newblock ``Deep residual learning for image recognition,''
\newblock in {\em CVPR}, 2016, pp. 770--778.

\bibitem{Everingham:voc:2015}
M.~Everingham, S.~M.~A. Eslami, L.~Van~Gool, C.~K.~I. Williams, J.~Winn, and
  A.~Zisserman,
\newblock ``The pascal visual object classes challenge: A retrospective,''
\newblock {\em International Journal of Computer Vision}, vol. 111, no. 1, pp.
  98--136, Jan. 2015.

\bibitem{Bochkovskiy:YOLOv4:2020}
Alexey Bochkovskiy, Chien-Yao Wang, and Hong-Yuan~Mark Liao,
\newblock ``{YOLOv4: Optimal} speed and accuracy of object detection,''
\newblock {\em ArXiv}, vol. abs/2004.10934, 2020.

\end{thebibliography}

\end{document}